# WHEN SHOULD A DECISION MAKER IGNORE THE ADVICE OF A DECISION AID?

Paul E. Lehner

School of Information Technology
George Mason University
4400 University Drive
Fairfax, Virginia   22030

Decision Science Consortium
1895 Preston White Drive
Reston, Virginia   22091

Theresa M. Mullin and Marvin S. Cohen

Decision Science Consortium

**Abstract**--This paper provides a simple probability analysis of the usefulness of decision aids that generate fallible advice. The main results are as follows.  The usefulness of a decision aid depends heavily on the ability of the user to identify contexts in which the aid (or user unaided) is likely to be incorrect. Without this ability, attending to (i.e., consider, but not always accept) the advice of a decision aid is counterproductive -- the decision maker would be better off either routinely accepting or routinely ignoring the aid's advice.

## 1.0   INTRODUCTION

In recent years, there has been a growing interest in the development of computer-based systems that provide analytical support to human decision making. The idea of using computers to provide decision making advice seems to have emerged somewhat independently from research in three separate areas:  information systems and optimization theory, from which the concept of a **decision support system** (DSS) seems to have emerged (e.g., Sprague and Carlson, 1982); psychological and mathematical research in judgment and decision (JDM) from which the idea of a **decision aid** emerged (e.g., Adelman, et.al, 1982); and artificial intelligence, which spawned the popular notion of an **expert system** (e.g., Hayes-Roth, et.al., 1983).

Common to these aiding traditions is the use of analytic techniques that the user is not otherwise likely to employ or have available to recommend possible solutions to a decision problem.  This appears to reflect a common underlying assumption that in contrast to computer systems that **only** provide problem-relevant **information**, a computer system that **also** provides good decision making **advice**, based on its own analysis, is more useful.  In other words, there is **added value** in performing a separate analysis and on that basis recommending solutions.

In this paper we critically examine this added-value assumption. Using a simple probability analysis, we demonstrate that it is unwarranted **as an assumption**.   Unless special measures are taken, it may be more reasonable to assume the opposite: that there is little value added by automatically generating advice. Conditions under which this result can be reversed are then presented.

In the remainder of this paper, we will use the term "decision aid" to refer all three types of aiding systems.

## 2.0 FALLIBLE VS. INFALLIBLE ADVICE

Modern technology has provided humankind with many computer-driven devices that perform so consistently that they are considered virtually infallible. That is, unless something quite unusual occurs, the device is simply trusted to perform its function. In some cases, this even applies to systems that automate decision making functions. For instance, as long as conditions are right (e.g., good weather), a good pilot will feel comfortable in relinquishing control to an automatic pilot.

By contrast, decision aids that recommend problem solutions are generally quite fallible. Indeed, an infallible advisory system is almost a contradiction in terms.

Unfortunately, there appear to be some inherent costs associated with fallibility. To illustrate these costs, consider the following scenario. A decision maker is given a decision aid that is based on an algorithm with an accuracy rate of 70%; that is, 70% of its recommended decisions are, by some accepted criteria, correct. The decision maker when unaided has an accuracy rate of 60%. Should the decision maker attend to the aids advice? The answer, of course, depends on how that decision maker chooses to use the aid. Let us assume further that **since the aid is known to be fallible**, the decision maker wishes to use some discretion in heeding the aid's advice. The user first considers the aid's advice and after some deliberation, will accept or reject the advice. We assume that the user accepts the aid's advice half the time. Finally if the user rejects the aid's advice, then he or she must solve the problem in the time remaining; with a reduced accuracy rate of, say, 40%.

Should our decision maker attend to the aids' advice? Clearly not! As shown in Figure 1, the probability of generating the correct answer drops to .55, even though the aid is based on an algorithm that performs significantly better than the unaided user!

More generally, using eq. 1 in Figure 1, one could characterize the comparison between unaided and aided performance as shown in Figure 2. As noted there, performance increases linearly with the probability of accepting an aid's advice. The usefulness of the advisory component will vary with changes in individual parameter values, but is in general maximal when the aid's advice is accepted **routinely**. In short, we get a paradoxical result: using a decision aid **as an aid** appears to be counterproductive.

Since this result may be counterintuitive, let us reconsider the plausibility of this scenario and examine possible

```
P(correct|aided) =

   P(advice correct|accept advice)*P(accept aid advice)
 + P(user correct|reject aid advice)*P(reject aid advice)

 = .7*.5 + .4*.5  = .55,
```

**FIGURE 1**
**Accuracy of Decision Maker/Decision Aid Combination: Equation [1]**
    [P(a|b) is the conditional probability of a given b.]

objections to eq 1. First it may seem implausible that the decision maker would excercise much discretion in accepting an aid's advice when the aid has a better hit rate. It seems unlikely, however, that a decision maker would simply transfer decision making responsibility to a fallible algorithm even **if** the user were aware of the algorithm's superior hit rate (Dawes, 1979).

Second, one might object that the process of considering and rejecting an aid's advice does not simply "use up" some of the user's problem-solving time. Other compensating benefits are provided. For example, examining an expert system rule trace might give a user partial analysis that will facilitate later problem solving even if the advice is rejected. We suggest, however, that although some decision aids **may** offer such incidental benefits, there is no reason to **assume** that most decision aids will. Also, even if there are such problem solving benefits, these benefits are not necessarily attributable to the advisory component of the aid. Partial analyses and problem relevant information can always be presented without also recommending solutions. One could argue that the user would be just as well off if he or she used the aid as an information source, but routinely ignored the aid's advice. (Actually the issue is bit deeper than this. Partial analyses can also be viewed as advice, leading to a recursive arguement. If certain classes of partial analyses are viewed as infallible (i.e., always trusted) then, of course, the user should simply accept the partial analysis results. Partial analyses that are viewed as fallible are subject to the same quantitative analysis presented here.)

Third, one might object to the assumption, implicit in eq. 1, that the user is unable to discriminate correct and incorrect advice. Thus, the probability of accepting the aid's advice is independent of whether or not the aid generated good advice. At first, this assumption might seem unreasonable; surely the user is more likely to accept good rather than bad advice. We suggest, however, that **unless** one has strong reasons to believe otherwise, this is what should be assumed. To suggest otherwise is to claim that the less accurate problem solver, prior to any independent problem solving, is able to outperform the more accurate problem solver. (This issue is investigated further in Section 3.0.)

Finally, it should be noted

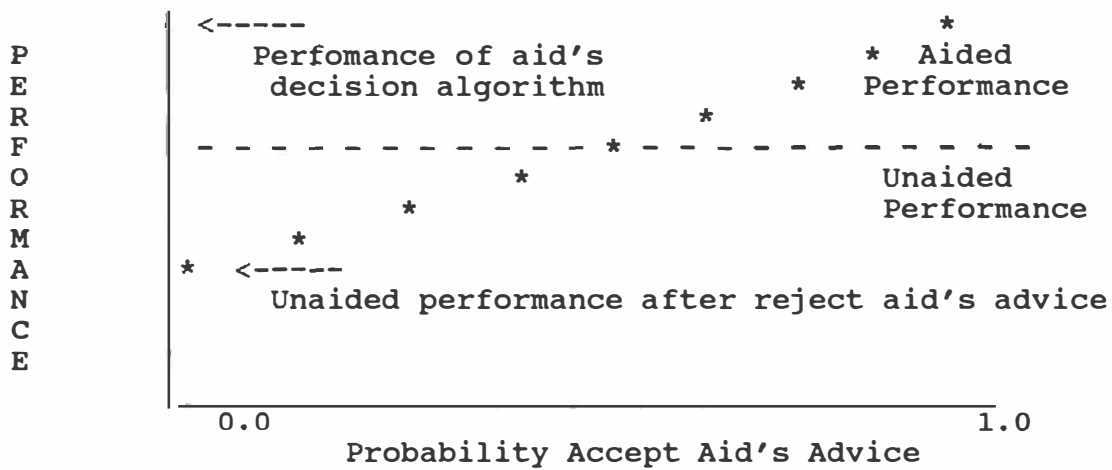

**FIGURE 2**
**Possible Relationship Between Aided and Unaided Performance**

that, for simplicity, solutions are treated as either correct or incorrect. This could be generalized to domains where there are, by some criteria, "better" and "worse" solutions. The analyses in this paper would then remain substantially unchanged, whether or not the criteria for discriminating good from bad solutions is explicitly known.

In summary, this analysis suggest two results. First, that a decision aid may be most beneficial if it is used as an automated decision maker, rather than as an aid. Second, if the user wishes to be careful about when to accept the aid's advice, then aided performance may actually be worse than unaided performance. Unfortunately, these results seem to contradict the spirit of decision aiding -- to build a useful interactive advisory system around a good, but fallible, algorithm.

### 3.0 OVERCOMING THE COST OF FALLIBILITY

Common sense suggests that "two heads are better than one." Despite the analysis of the previous section, intuition suggests that this maxim should applicable to combined decision maker/decision aid problem solving. Furthermore, it is clear that the **potential** performance of a decision maker/decision aid combination is higher than either problem solver independently. In the example of the last section, for instance, if the aid and unaided user are conditionally independent in their hit rates, then the potential combined hit rate is .88 -- the probability that at least one of the two problem solvers will generate the correct answer. Of concern, therefore, is the determination of the conditions under which this potential can be realized.

Of course, improvements that change any of the various parameter values in eq. 1 will help. But such changes will not change the result that performance is maximal when the aid is used as an automated decision maker. Furthermore, improvements in the incidental benefits obtained from the process of evaluating, but rejecting, the aid's advice will decrease the added-value uniquely attributable to the advisory component.

We suggest instead that a circumstance where the user/decision aid combination performs better than either problem solver independently is only realizable if the decision maker is discriminating about when to accept the aid's advice. To illustrate why, consider the modification of eq. 1 shown in Figure 3. In the example in Section 2.0, we assumed that;

$P(\text{accept advice} | \text{advice correct}) = .5$
$P(\text{accept advice} | \text{advice wrong}) = .5$
$P(\text{user correct} | \text{reject advice \& advice correct}) = .4$
$P(\text{user correct} | \text{reject advice \& advice wrong}) = .4$
$P(\text{advice correct}) = .7$,

which resulted in $P(\text{correct} | \text{aided}) = .55$.

Consider now a case where the user is somewhat discriminating about when to accept/reject the aid's advice. For instance,

$P(\text{accept advice} | \text{advice correct}) = .7$
$P(\text{accept advice} | \text{advice incorrect}) = .3$.

Now $P(\text{correct} | \text{aided}) = .66$. A moderate ability to discriminate good from bad advice resulted in a moderate improvement in performance, more than compensating for the cost of evaluating and rejecting the aid's advice. (Note that some of the improvement is attributable to the fact that the marginal P(accept advice) has also increased. This is reasonable -- if the aid is usually correct, then a discriminating user would usually accept its advice.)

A more dramatic example is where the user is very discriminating about accepting the aid's advice, but the aid's hit rate is lower than the user. For example,

$P(\text{accept advice} | \text{advice correct}) = .9$
$P(\text{accept advice} | \text{advice wrong}) = .1$
$P(\text{advice correct}) = .55$.

We now get $P(\text{correct} | \text{aided}) = .68$.

Another variant is the case where the user is a good predictor of his or her own capabilities, but does not have a good assessment of the aid's. This suggests a scenario where the user must first problem whether to solve the problem unaided. If yes, then ignore the aid. If not, then accept the aid's advice. This variant leads to the equation shown in Figure 4.

Assume that the user can predict his or her own success 70% of the time, the probability that the aid is correct is independent of whether or not the user would be correct, and (for simplicity) that it takes no time to decide whether to use the aid. We can then plug in the following values:

$P(\text{correct} | \text{aided}) = .7*.6 + .7*.3*.6 + .7*.7*.4 = .742$.

Finally, users who are discriminating in both ways would

$$P(\text{correct}|\text{aided}) =$$

$$P(\text{accept advice}|\text{advice correct})*P(\text{advice correct})$$
$$+ P(\text{user correct}|\text{reject advice \& advice correct})$$
$$\quad *P(\text{reject advice}|\text{advice correct})$$
$$\quad *P(\text{advice correct})$$
$$+ P(\text{user correct}|\text{reject advice \& advice wrong})$$
$$\quad *P(\text{reject advice}|\text{advice wrong})*P(\text{advice wrong}).$$

**FIGURE 3**
**Accuracy of Decision Maker/Decision Aid Combination: Equation [2]**

likely perform even better.

As these quantitative examples suggest, significant benefits can be obtained by supporting a user's ability to be discriminating in his or her use of an aid. This suggests that a user must know enough about an aid's falliblities (or her own) to be able to identify circumstances when the aid's advice (or her own judgment) should simply be ignored. Unfortunately, and not suprisingly, few decision aids are designed with the intent of helping a user easily identify contexts in which the aid is likely to be **incorrect**. Similarly, developers intent on transfering an aid to an operational environment are not likely to advertise the weaknesses of their product. Consequently, there is little reason to presume that the advisory component of many decision aids contribute much in the way of decision **support** (vs. automation).

Fortunately, however, there seem to be a variety of relatively simple ways to achieve this discrimination. The first, and most obvious, is simply to promote an accurate mental model of the decision aid in the decision maker. In Lehner and Zirk (1987), for instance, it was found that even a rudimentary understanding of how an expert system works could lead to dramatic improvement in performance. Alternatively, one could embed within the decision aid itself some "metarules" for identifying contexts in which the aids advice should probably be ignored. For instance, an aid based on a quantitative uncertainty calculus might be able to flag situations where there is significant higher order uncertainty (significant amount of missing data, sensitivity analysis indicates recommendation not robust, etc.). Finally, one could promote, in decision makers, a better understanding of the human decision making process; perhaps making them aware of many of the common biases (e.g., hindsight bias) that lead decision makers to be overconfident in their assessment accuracy.

Before closing this section, it is worth noting that in eqs. 1 - 3 we assumed that after rejecting the aid's advice the probability of generating the correct answer unaided is the same whether or not the rejected advice was correct. Presumably, however, the decision maker and the aid's algorithm are based on a common source of knowledge. One might

```
P(correct|aided) =

  P(ignore aid|user correct)*P(user correct)
+ P(advice correct|uses aid)*P(uses aid|user correct)
                            *P(user correct)
+ P(advice correct|uses aid)*P(uses aid|user wrong)
                            *P(user wrong)
```

**FIGURE 4**
**Accuracy of Decision Maker/Decision Aid Combination: Equation [3]**
("user correct" could be expanded to "user would be correct".)

therefore expect a positive correlation between the two problem solvers. If, however, the two problem-solving approaches are positively correlated (e.g., when an expert system is designed to mimic human problem solving), then the aided probability of success can easily **decrease**. To illustrate this, recall the example at the beginning of Section 3.0 where we had

  P(accept advice|
      advice correct) = .7
  P(accept advice|
      advice incorrect) = .3.

which resulted in P(correct|aided) = .68.

Now suppose we added the assumption that the aid's algorithm is uniformly better than the unaided user. For all problems where the unaided user would generate the correct answer, the algorithm would also get it right. We then add to our scenario:

  P(user correct|reject advice
      & advice correct) = .6/.7
  P(user correct|reject advice
      & advice wrong) = 0.

Plugging these new numbers into eq. 2 gives us P(correct|aided) = .67.

The effect of the dependency was small, but negative. More generally, P(correct|aided) decreases whenever the user is more likely to generate a correct answer in the same circumstances as the aid. For the classes of problems we explored the effect of dependencies of this type was usually small. The impact of such dependencies was therefore ignored in our analysis. However, this result does suggest that the popular notion that decision aids should be designed to mimic human expert problem solving may be misguided.

## 4.0 DISCUSSION

To summarize. The usefulness of a decision aid depends on thea user's ability to identify contexts in which the aid (or user unaided) is likely to be incorrect. Without this ability, "considering" the advice of a decision aid is counterproductive -- the decision maker would be better off either routinely accepting or ignoring the aid's advice. The same result holds for any partial analyses.

We do not claim that the simple probability analysis presented here is a realistic model of all the subtleties of a user/decision aid interaction. We do, however, claim that it provides a reasonable charac-terization of the impact of the variables we are examining. It is hard to imagine how a more complex model would would suggest directional impacts different from those presented in this paper.

Also of interest is the relationship of this analysis to empirical research examining the effectiveness of decision aids (see for reviews Adelman, in press; Sharda, et.al., 1988). Empirical tests have had mixed results -- some aids improve performance, others have little effect, and some decrease performance. Unfortunately, most empirical efforts to evaluate a decision aid evaluate the aid as a whole. They do not attempt to discriminate the contribution of various components of an aid. In this paper we have examined the impact of various components analytically. Our analysis does not suggest that decision aids per se are ineffective, but only that it is inappropriate to attribute effectiveness to advisory support provided by the aid. While a particular decision aid may be useful, that usefulness could be attributed to the fact that the aid also serves as an information system, and may also generate advice that is routinely accepted. It should not be assumed that the advisory component, which is the core of most decision aids, provides any useful decision support. Empirical research in this area is needed.

## ACKNOWLEDGEMENTS

We would like to thank Len Adelman, Rex Brown and Dave Schum for their comments on an earlier draft of this paper.